%% file: weirules-Pitsikalis.tex
\documentclass[runningheads]{llncs}
\usepackage{graphicx}
\usepackage{booktabs}
\usepackage{pgfplots}
\usepackage{xcolor}
\usepackage{amssymb}
\usepackage{amsmath}
\usepackage{listings}
\usepackage{multirow}
\usepackage{comment}
\usepackage{float}
\usepackage{subfig}
\usepackage{pgfplots}
\pgfplotsset{compat=newest}
\usepackage{siunitx}
\usepackage{diagbox}
\usepackage{cite}
\usepackage{comment}

\newcommand{\bblue}[1]{\textcolor{blue}{\textbf{#1}}}
\newcommand{\bred}[1]{\textcolor{red}{\textbf{#1}}}

\makeatletter
\newcommand{\Vast}{\bBigg@{11}}
\makeatother

\makeatletter
\newcommand\makebig[2]{%
  \@xp\newcommand\@xp*\csname#1\endcsname{\bBigg@{#2}}%
  \@xp\newcommand\@xp*\csname#1l\endcsname{\@xp\mathopen\csname#1\endcsname}%
  \@xp\newcommand\@xp*\csname#1r\endcsname{\@xp\mathclose\csname#1\endcsname}%
}
\makebig{Biggg} {4.5}

\newenvironment{nalign}{
    \setlength\abovedisplayskip{0pt}
    \begin{equation}
    \begin{aligned}
}{
    \end{aligned}
    \end{equation}
    \ignorespacesafterend
}

\definecolor{mygray}{gray}{0.6}

\begin{document}

\title{Logic Rules Meet Deep Learning: A Novel Approach for Ship Type Classification}
\titlerunning{Logic Rules Meet Deep Learning}

\author{ Manolis Pitsikalis\inst{1} \orcidID{0000-0003-2959-2022} \and Thanh-Toan Do \inst{2} \orcidID{0000-0002-6249-0848} \and Alexei Lisitsa\inst{1} \orcidID{0000-0002-3820-643X}  \and Shan Luo\inst{1} \orcidID{0000-0003-4760-0372}}

\authorrunning{M. Pitsikalis et al.}

\institute{Department of Computer Science, University of Liverpool
\\ \email{\{e.pitsikalis,a.lisitsa,shan.luo\}@liverpool.ac.uk} 
\and Department of Data Science and AI, Monash University\\ \email{toan.do@monash.edu}\\
}

\maketitle 
\input{abstract}

\input{introduction}
\input{relatedwork}
\input{background}
\input{method}
\input{evaluation}
\input{conclusion}

\section*{Acknowledgements}
This work has been funded by the Engineering and Physical Sciences Research Council (EPSRC) Centre for Doctoral Training in Distributed Algorithms at the University of Liverpool, and Denbridge Marine Limited\footnote{https://www.denbridgemarine.com}, United Kingdom.
%
%
%
\bibliographystyle{splncs04}
\bibliography{bibliography}
\end{document}

%% file: abstract.tex
\begin{abstract}
The shipping industry is an important component of the global trade and economy, however in order to ensure law compliance and safety it needs to be monitored.  In this paper, we present a novel Ship Type classification model that combines vessel transmitted data from the Automatic Identification System, with vessel imagery. The main components of our approach are the Faster R-CNN Deep Neural Network and a Neuro-Fuzzy system with IF-THEN rules. We evaluate our model using real world data and showcase the advantages of this combination while also compare it with other methods. Results show that our model can increase prediction scores by up to 15.4\% when compared with the next best model we considered, while also maintaining a level of explainability as opposed to common black box approaches.
\keywords{Object detection  \and Classification rules \and Fuzzy rules}
\end{abstract}

%% file: introduction.tex
\section{Introduction}
Nowadays, the combination of deep learning with logic has been attracting a lot of attention for numerous reasons. On the one hand, deep learning has been used extensively in different applications, such as object detection~\cite{ren_faster_2015,redmon2015look} and language analytics tasks~\cite{Mikolov_2013,Collobert_2008} with a lot of success. On the other hand, logic approaches have been used widely in tasks where explainability is required, such is the case in certain medical tasks~\cite{London_2019}, or where expert knowledge is available and needs to be encoded into a model~\cite{Grosan2011}. 
However, logic approaches typically provide crisp predictions over symbolic data, and it is often the case that human experts are required to express the knowledge into some form of logic. On the other hand, although deep learning approaches can handle unstructured data such as text and images, their black box nature makes them inadequate for tasks where explainability is a key requirement.
For these reasons, there is a need for a model that combines the advantages of deep learning with those of logic based approaches. In this paper we propose a model that combines deep learning with logic rules, applied in the shipping domain, for the task of ship type classification.

Shipping, since the ancient years, has been a very important component of global trade and economy. However, there are many cases where ships are found to be involved in illegal activities that are possibly dangerous or harmful to the environment. In order to ensure law compliance and safety, shipping needs to be monitored. Today, there is abundance of Maritime data. Systems such as the Automatic Identification System (AIS\footnote{http://www.imo.org/en/OurWork/Safety/Navigation/Pages/AIS.aspx}), a system that allows the transmission of both dynamic spatio-temporal data and static identity data from vessels, CCTV imagery from stationary cameras placed in ports or from cameras in Drones, Satellite Aperture Radar (SAR) images, data from RADAR systems and so on,  provide valuable information for the maritime monitoring task. Ships or in general maritime vessels are divided into ship types based on their characteristics and purpose, for example there are fishing vessels, search and rescue vessels, cargo vessels and so on, consequently different regulations apply to each ship type. In order to promote security and abidance to regulations the process of classifying and validating a vessel's type needs to be automated using the data available.

\begin{figure}[htpb]
\centering
\includegraphics[width=.9\columnwidth]{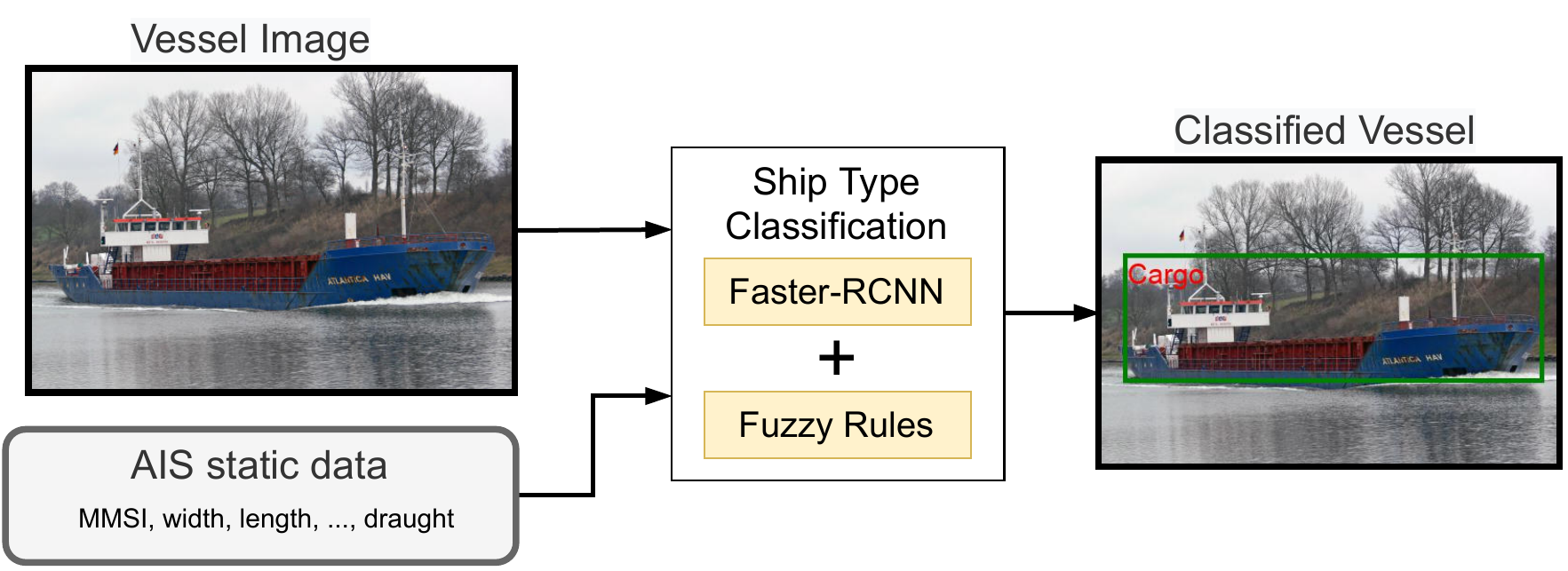}
\caption{Pipeline of the Ship Type Classification system.}
\label{fig:pipeline}
\end{figure}

In the context of this paper, we aim to combine two data sources in order to perform ship type classification as illustrated in Figure~\ref{fig:pipeline}.
We use static  vessel transmitted AIS data and a collection of images pre-linked with the AIS records. The main components of our approach are the Faster-RCNN deep neural network and a  Neuro-Fuzzy model leveraging convolutional deep features of the former along with AIS information. We evaluate the presented model using real world data and compare it with other algorithms available. The contributions of this paper are:
\begin{itemize}
\item a novel classification model that combines images with structured contextual data, in our case vessel images and AIS data for ship type classification,
\item a Neuro-Fuzzy approach that can be used to improve classification accuracy in object detection and maintain a level of explainability when additional data is available.
\end{itemize}

The paper is organised as follows. In Section~\ref{sec:related} we present work related to our approach. In Section~\ref{sec:bgr} we present material needed for the presentation of our work. Next, in Section~\ref{sec:methodology} we describe our methodology, while in  Section~\ref{sec:eval} we present our experimental setting and analyse the experiments we performed in our evaluation. Finally, in Section \ref{sec:sum} we focus on future work and discuss our findings.

%% file: relatedwork.tex
\section{Related Work}
\label{sec:related}
In this section we present works related to the ship type classification task, the combination of logic rules and neural networks and finally the fuzzification of logic rules.

In the recent years, Ship Type classification has attracted a lot of interest. D. Nguyen et al. in~\cite{Nguyen2018} present an architecture capable of performing ship type classification and anomaly detection using AIS trajectories. They use an embedding block based on Variational Recurrent Neural Networks (VRNNs)~\cite{Chung2015} to transform the irregularly shaped AIS data into 10-min sampled structured regimes. Then using the regimes of the embedding block, they perform specific tasks such as trajectory reconstruction, ship type classification, anomaly detection etc. In our approach, instead of using complete trajectories, we choose to use only the information transmitted over the static AIS messages, since complete trajectories are not always available and require a bigger amount of transmitted AIS messages over a period of time.
U. Kanjir et al, in another work~\cite{kanjir2018} provide an overview of the available methodologies for vessel detection and classification using SAR imagery along with the most important factors affecting the accuracy of the aforementioned task. According to their overview, the ship detection task comprises of three steps. The first step is to detect whether a vessel exists in an image; the second step is to identify its type, while the third step is the unique identification of the vessel, i.e., finding a unique identifier such as the Maritime Mobile Service Identity (MMSI) number or the International Maritime Organisation (IMO) number. Their analysis focuses on the optical sensors used, the different detection workflows used for detecting vessels in images, the classification methods used, the metrics for evaluation and finally on the importance of combining different types of data sources for better results. We focus on the two first steps of U. Kanjir's et al. ship detection methodology, which are the `detection of a ship in an image' and  `the identification of its type'.

When it comes to Logic in Neural Networks, Z. Hu et al. in \cite{hu2016},  present a way of harnessing knowledge from logic rules in a teacher-student model setting. The goal of their approach is to transfer the human knowledge  provided by the rules into the network parameters. They evaluate their rule-knowledge distillation approach in certain Natural Language Analytics tasks and their results show that their approach manages to improve the accuracy compared to other methods. However this approach does not allow any learning in the aspect of the logic rules, since they remain constant during training.
In another work, G. Marra et al.~\cite{Marra2020}, integrate first order logic rules with deep learning in both training and evaluation settings. The approach of G. Marra et al. allows the learning of the weights of the neural network and the parameters of the integrated logic rules that are used for reasoning. In order to render the rules suitable for integration they replace conjunction and disjunction with the t-norm and s-norm equivalents. 
In our approach, we fuzzify the logic rules, with the use of the sigmoidal membership function and use weighted exponential means for the approximation of conjunction and disjunction. The fuzzified rules are then integrated into the neural network architecture, whereby the weights of the disjunction operation are learned. 

M. Tsipouras et al. in~\cite{tsipouras_automated_2008} present a fuzzy model created using rules extracted from a C4.5 Decision Tree that they later fuzzify  in a similar way as we do in this paper. They evaluate their optimised fuzzy model on medical data and their results show that they achieve comparable accuracy with an ANN approach. Compared to our approach, instead of using one Decision Tree to extract the logic rules, we use a set of one-vs-all CART classifiers since this method produces specialized rules for each class~\cite{Hashemi_Sattar_2009} in our problem. Moreover, we treat our  fuzzy model as a way of integrating two sources of information, in the training step the initial fuzzy model is created using one source of information while the parameters of the fuzzy model are computed using both sources, then in the evaluation step both sources of information are required to make a prediction.

%% file: background.tex
\section{Preliminaries}
\label{sec:bgr}
The main components of our approach are a Fuzzy model created by extracting rules from a set of Decision Trees, and the Faster R-CNN deep neural network~\cite{ren_faster_2015}. In what follows, we present the necessary background for our approach.

Decision Tree classifiers are probably the most widely known classifiers in Machine Learning. In their simplest form, they use a binary tree structure, that when traversed in accordance to a part of an input set they produce a conclusion.
In our approach we use the Classification and Regression Trees (CART)~\cite{reason:BreFriOlsSto84a}. The tree growing process of CART tree starts from the root node, containing all the data instances, and recursively continues to its child nodes by choosing at each node the split among all possible univariate splits that makes the instances of its child nodes the purest. Common splitting criteria include the Gini criterion and the Twoing criterion. A node stops splitting when one of the following conditions is satisfied. The node is pure, i.e., all its instances have the same label; the values of the instances in the node are the same; the depth  has reached a user specified depth; the number of the instances in the node are less than a user specified threshold or if its splits results in a child node  containing a number of instances less than a user defined threshold. Thus, the tree growing process is completed when no further splits can be performed.

Fuzzy Logic as opposed to Boolean Logic produces values ranging from 0 to 1 instead of crisp 0 and 1. However, crisp values can be converted to fuzzy values with the use of fuzzy membership functions. For example, a fuzzy membership function for the comparison of two numbers such as $a > b$ that transforms the True (1) or False (0) result to a value $\in(0,1)$ can be the sigmoidal membership function $\mathit{f}(a;b,s)$ with a center at $b$, a slope parameter $s$ and the value of $a$. Moreover, the logical connectives of conjuction and disjunction in a Logic Formula can be replaced with their fuzzy equivalents such as $\min$ and $\max$ respectively.

Faster R-CNN~\cite{ren_faster_2015} is a deep neural network that is used for object detection, that is, the process of identifying and locating objects in images or videos. The architecture of Faster R-CNN consists of a Convolutional Network for feature extraction--usually a pre-trained version of well known image classification network such as ResNet50~\cite{he2015deep}, a Regional Proposal Network (RPN) that is used for the generation of Regions of Interest (RoIs),  and finally a Classification and Regression Network that takes input convolutional features from a RoI pooling layer applied on the features from the Convolutional Network and the ROIs. In our approach we extract convolutional features from the output of the RoI pooling layer of a pretrained Faster R-CNN network corresponding to the detected objects in the input images.

%% file: method.tex
\section{Methodology}
Here, we present our methodology of combining the information of two data sources of different format---the first source contains images while the second one contains numerical structured data describing attributes of vessels---in order to achieve higher prediction scores and add explainability to the final model. 
\label{sec:methodology}
\subsection{Rule Extraction and Fuzzification}
\label{sec:rule-ex}

As mentioned before the first stage of our method is the extraction of logic rules in Disjunctional Normal Form (DNF). For each class label $y$, we train a Decision Tree model using the CART algorithm for binary classification where the positive class is $y$ and the remaining are negative. Then, for each class $y$ we parse the corresponding $tree$ and recursively create a rule, by adding conditions expressing the path from the root to leaf nodes where the label is $y$. Therefore, a condition $C_i$ included in the body of the rule concerning label $y$ is expressed as:
\begin{center}
\begin{nalign}
C_i =(x_1\ \mathit{op}\ v_1 )\wedge\cdots\wedge(x_k\ \mathit{op}\ v_k)
\end{nalign}
\end{center}
where $v_i$ are the values obtained during the splitting process of the tree training; $x_i$ are the values of the attributes on which the constraints are applied and $\mathit{op}$ is either `$>$' or `$\leq$'. A rule $R_i$ for a specific class $y_i$ has the following form:
\begin{center}
\begin{nalign}
&\mathit{IF}\ C_{1}\vee \cdots \vee C_{n}\ \mathit{THEN}\ y_{i}
\end{nalign}
\end{center}

\begin{figure}[ht]
\centering
\resizebox{0.4\columnwidth}{!}{
\begin{tikzpicture}
\Large
    \begin{axis}[xmin=-5,xmax=5,ymax=1.05,ymin=-0.05,samples=50,grid=major,xlabel={$x$},ylabel={$c_{i,\{>,<\}}'$},title={$s=\{1,3\}, v=0$}]
    \addplot[blue]({x},{1/(1+e^(-1*(x-0))});
    \addplot[blue, dashed]({x},{1/(1+e^(-3*(x-0))});
    \addplot[red]({x},{1/(1+e^(-1*(-x+0))});
    \addplot[red, dashed]({x},{1/(1+e^(-3*(-x+0))});
\end{axis}
\end{tikzpicture}
}
\caption{Plots of $c_{i,>}'$ (blue) and $c_{i,<}'$ (red) with $s$ set to 1 and 3 (continuous and dashed lines respectively)  for $x \in [-5,5]$.}
\label{fig:sigfmplots}
\end{figure}
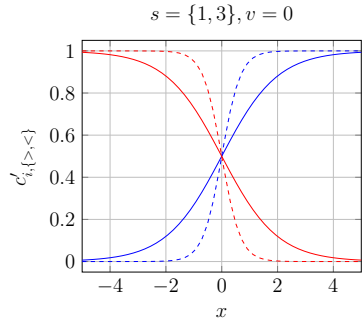

For the Fuzzification of the rules, we apply the sigmoidal membership function~\eqref{eq:sigmf} in the comparisons $c_{ij} \in C_i$ so that each comparison $c'_{ij}$ yields a value in $[0,1]$. 
\begin{center}
\begin{nalign}
\label{eq:sigmf}
&f_>(x;s,v)=\frac{1}{1+e^{-s(x-v)}}\ \mathrm{for}\ x > v\  \mathrm{and},&\\
&f_{\leq}(x;s,v)=\frac{1}{1+e^{-s(v-x)}}\ \mathrm{for}\ x \leq v&
\end{nalign}
\end{center}
The $s$ parameter defines the slope of the sigmoid curve while the $v$ parameter defines the `center' of the curve (see Figure~\ref{fig:sigfmplots}). 
Moreover, as seen in equation~\eqref{eq:weighted_rule}, for each class rule $R_i$ we add a weight $w_i$ in each of the conditions $C_i$ where $w_i \in [0,1]$ and $\sum\limits_{i=0}^n w_i =1$.
\begin{center}
\begin{nalign}
\label{eq:weighted_rule}
&\mathit{IF}\ w_1 C_{1}\vee \cdots \vee w_n C_{n}\ \mathit{THEN}\ y_{i}
\end{nalign}
\end{center}
Finally, we replace the logical connectives of conjunction and disjunction with their Weighted Exponential Mean approximation~\cite{dujmovic_generalized_2007} as depicted in Table~\ref{tab:minmax_approx}, $\min'$ and $\max'$ respectively. For the experimental results presented in Section~\ref{sec:eval} we have set the level of andness and orness to `medium high' $r =-5.4$ for conjunctions and $r=5.4$ for disjunction by taking into consideration the results of the ablation study presented in the same section.
\begin{table}[ht]
\centering
\caption{Approximation of conjunction and disjunction using Weighted Exponential Means. An analysis of the effects of $r$ for WEM  can be found in~\cite{dujmovic_generalized_2007}.}
\label{tab:minmax_approx}
\setlength{\tabcolsep}{8pt}
\begin{tabular}{ll}
\toprule
 Notation  &  WEM\\\midrule
$ c_1\wedge \cdots \wedge c_n$ &          $\frac{1}{r}\ln\left(\frac{1}{n}\sum\limits_{i=0}^n e^{r c_i}\right),r \in\{-14.0,-5.4, -2.14\}$\\[10pt]
 $ w_1 C_1 \cdots \vee w_n C_n$ &       $\frac{1}{r}\ln\left(\sum\limits_{i=0}^n w_i e^{r C_i}\right),r\in\{2.14,5.4,14.0\}$\\\bottomrule
\end{tabular}
\end{table}    
Consequently, the truth value of a rule $R_i$ is produced by computing the $\max'$ of all the conditions $C_i$ as  follows:
\begin{center}
\begin{nalign}
R_i(x;W,S)={\max}'\bigg\lbrace w_1{\min}'\lbrace c'_{11},\cdots, c'_{1k}\rbrace,\dots, w_n{\min}'\lbrace c'_{n1},\cdots, c'_{1z} \rbrace \bigg\rbrace
\end{nalign}
\end{center}
where $W$ is vector containing the weights of the conditions and $S$ is vector containing the slope parameters of the sigmoid membership functions included in each $C_i$. Finally, to make a class prediction, we produce  the $R_i$ values for each class $y_i$ and produce the simplex vector by applying $L1$ normalisation on the vector $F_c = \{R_1, \dots R_m\}$ where $m$ is the number of classes.

\subsection{Neuro-Fuzzy Combination}
\begin{figure*}[htpb]
\centering
\includegraphics[width=0.9\textwidth]{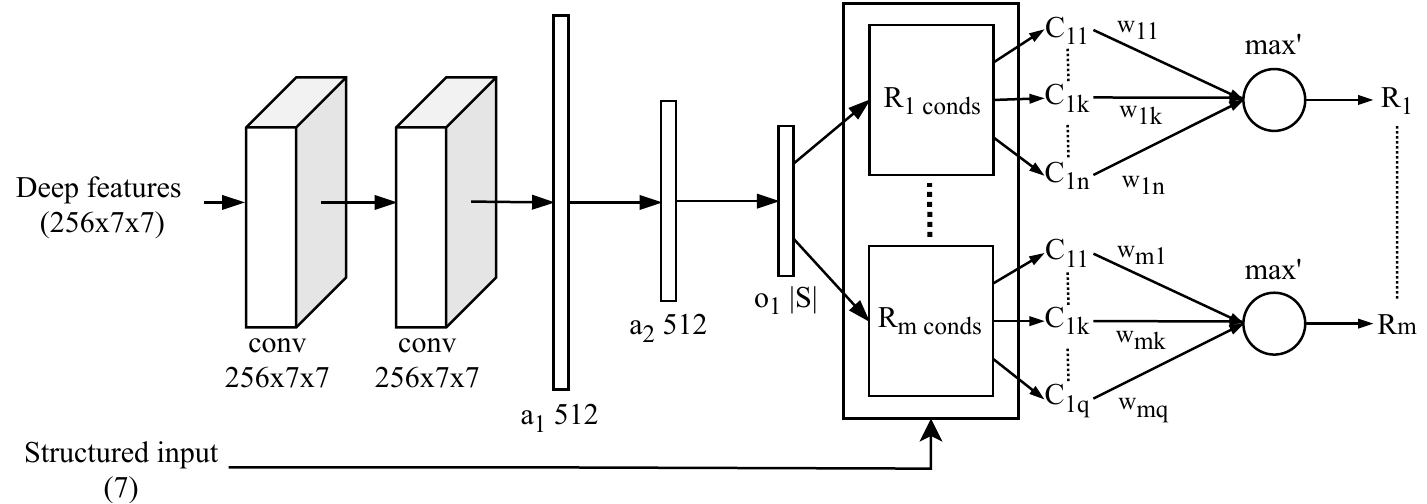}
\caption{Architecture of the Neuro-Fuzzy network. The upper branch receives deep features extracted from the RoI pooling layer of a Faster R-CNN model when given a vessel image, while the lower branch receives AIS data (Structured data) that is then given to the different fuzzified rules $R_{i\ \mathit{conds}}$. }
\label{fig:arch}
\end{figure*}

In the previous Section, we mentioned that each class rule $R_i$ accepts as parameters a set of weights $W$ for the weighted max approximation and a set of slopes $S$ for the sigmoid membership functions included in the conditions of the rule. Here, we present how we combine the Fuzzy Model described in Section~\ref{sec:rule-ex} with a Neural Network into a single model. The architecture of our combined model is illustrated in Figure~\ref{fig:arch}.

In our approach we aim to perform ship type classification by combining the information included in the images of the vessel with the information included in the characteristics of the vessel transmitted over AIS. For this reason, the architecture of our model has two inputs. The first one is the deep convolutional features extracted from the RoI pooling layer of a pre-trained Faster R-CNN model, while the second input includes the values of the AIS fields. In the upper branch of Figure~\ref{fig:arch} we use two convolutional layers, where the output of the second layer is flattened and given as input to a fully connected layer $a_1$ with  512 neurons followed by the fully connected layer $a_2$, with batch normalization, dropout and ReLu activation, and finally the $o_1$ layer which has Leaky ReLu as an activation function and yields an output equal to the size of $S$, i.e, the vector containing the slope parameters of the rules.

Then, using the output of the $o_1$ layer, along with the rules input (AIS fields) we can now compute the values $C_{ij}$ included in each $R_i, i\in[1,m]$  (see the $R_{i\  \mathit{conds}}$ blocks in Figure~\ref{fig:arch}). Next, for each $R_{i\ \mathit{conds}}$ we create the vector $\{e^{r C_{i1}},\dots, e^{r C_{in}} \}$ and feed it into a log activated layer, with bias set to 0 and normalised weights, that computes  the approximation of weighted disjunction as follows:
\begin{center}
\begin{nalign}
\label{eq:final-output}
R_i=\frac{1}{r}\ln\left(\sum\limits_{j=0}^n w_{ij} e^{r C_{ij}}\right)
\end{nalign}
\end{center}
where $w_{ij}$ are the weights of the input and $r$ is the orness level. 
Finally, all $R_i$ are fed through a \textit{softmax} layer that outputs a probabilistic vector $F$. 

We train the complete neuro-fuzzy model using the cross entropy loss $L$ (Equation~\eqref{eq:loss}) over $F$ and the one-hot ground truth label $y = \{y_c\}_1^m$.
\begin{center}
\begin{nalign}
\label{eq:loss}
L = -\sum_{c=1}^{m}y_c\log(f_c),\ F=\{f_1, \dots f_m\}
\end{nalign}
\end{center}

\subsection{Deep Feature Extraction}
For each vessel image we extract a deep convolutional feature from a pre-trained Faster R-CNN network in prediction mode. In detail we keep the $256\times7\times7$ feature vector from the output of the RoI pooling layer, corresponding to the bounding box that yields the highest confidence score after the non-maximum suppression stage. Using the collected deep convolutional features and the AIS data we proceed into training our neuro-fuzzy model.
\label{sec:extract}

%% file: evaluation.tex
\section{Evaluation}
In this section, we present the characteristics of the datasets we use for our experimental evaluation, the experimental settings for the training of our models and finally the prediction scores of the evaluated models. 
\label{sec:eval}
\subsection{Dataset}
We construct our dataset using the AIS records from the maritime dataset presented in  \cite{RAY2019104141}. In the context of the presented experiments, we use only the `MMSI', `to\_bow', `to\_stern', `to\_starboard', `to\_port', `width' , `length' and `draught' fields of the static AIS messages (see Table~\ref{tab:ais_fields} for a description of those fields). 
\begin{table}[ht]
\caption{Description of the retained AIS fields.}
\label{tab:ais_fields}
\centering
\small
\begin{tabular}{@{}ll@{}}
\toprule
AIS Field    & Description                                                            \\ \midrule
MMSI         & The Maritime Mobile Service Identity number                                \\[2.5pt]
to \{bow, stern, &  Distance from the AIS transceiver to the \{bow, stern, \\
starboard, port\} &      right side, left side\} of the vessel               \\[2.5pt]
draught      & The vertical distance between the waterline\\& and the bottom of the hull \\[2.5pt]
width        & The width of the vessel (to bow + to stern)                                                      \\[2.5pt]
length       & The length of the vessel (to starboard + to port)                                                 \\ \bottomrule
\end{tabular}
\end{table}
Moreover, using the `MMSI' value we collect up to 5 images for each vessel from the IHS Markit World Register of Ships (v12) and the ShipScape photographic library. The collected images contain one vessel per image and have been annotated using the ship type field of the AIS messages and the manual selection of the bounding box of each ship. Note that there are more images than distinct vessels, since each vessel may have up to 5 images.  Using the retained AIS fields and the collected images we create an Image classification centred dataset $\mathit{IC}$ by computing the following natural join:
\begin{center}
\begin{nalign}
I \underset{\text{\tiny I.MMSI=A.MMSI}}{\bowtie} A
\end{nalign}
\end{center}
where $I$ is the image table with the fields `Image\_ID', `MMSI' and `df' corresponding to the deep feature of the image while $A$ is the table with the mentioned AIS fields. 
However, there is also another way of looking at the classification problem. While in the previous case the problem is image centralised, in the current case we focus on the vessels, therefore we create a vessel centred dataset $\mathit{VC}$ by grouping and averaging the deep features per vessel MMSI as follows:
\begin{align}
\centering
A \underset{\text{\tiny A.MMSI=I.MMSI}}{\bowtie} \gamma_\text{\scriptsize MMSI, avg(df)}(I)
\label{eq:dt2}
\end{align}
The number of different vessels and images per vessel type are presented in Table~\ref{tab:comb_dataset_instances}.
\begin{table}[t]
\centering
\caption{Number of instances per class.}
\label{tab:comb_dataset_instances}
\begin{tabular}{lll}
\toprule
Shiptype  & Vessels & Images\\ \midrule
Cargo 	  & 		2412 & 11185\\
Tanker    & 		 864 & 3950\\
Other     & 		 53 & 229\\
Passenger &            42 & 199\\
Tug &            32 & 139\\\midrule
Total  &   3403 & 15702\\\bottomrule
\end{tabular}
\end{table}
Note that the numbers presented in Table~\ref{tab:comb_dataset_instances} correspond to records that have images available; records that are incomplete are not used in the experiments. Additionally, we remove all instances of ship types that do not have more than 20 different vessels.
\subsection{Baseline model}
\begin{figure}[t]
\centering
\includegraphics[width=0.6\columnwidth]{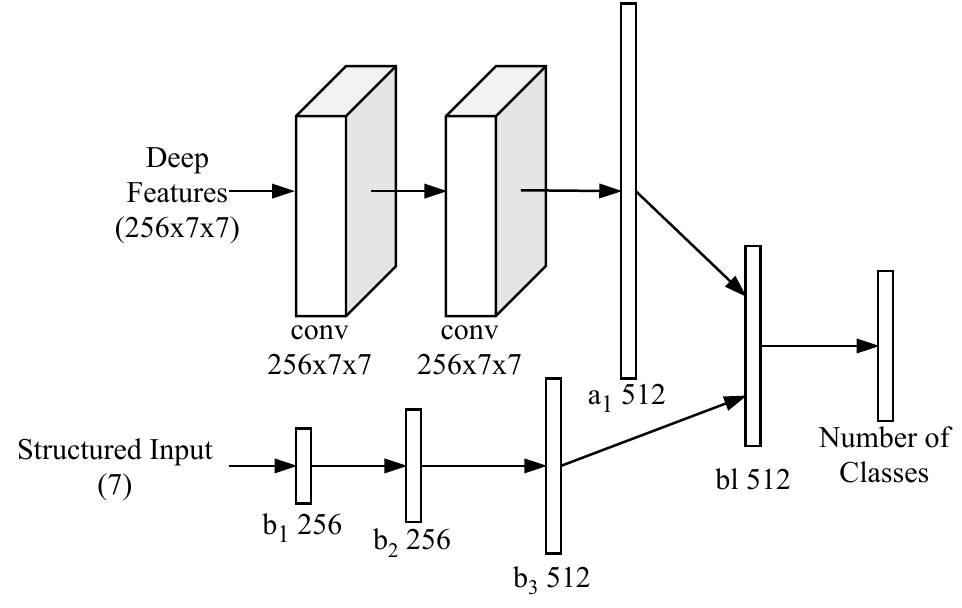}
\caption{Architecture of the baseline model. Here, in contrast with the model of Figure~\ref{fig:arch}, AIS data is given as input to the $b_1$ layer of the lower branch, while the combination of the different data sources is achieved using the bilinear layer $bl$.}
\label{fig:baseline}
\end{figure}

In addition to the Neuro-Fuzzy model presented in this paper, we create the baseline model of Figure~\ref{fig:baseline} which  retains the convolutional branch of the neuro-fuzzy model up to layer $a_1$ and adds a second branch that accepts as input the AIS fields. The additional branch has one input layer with 7 input neurons and 256 output neurons,  followed by two  fully connected layers ($b_2, b_3$) with batch normalization, dropout and ReLu activation. The output of layers $a_1$ and $b_3$ is then given as input to the bilinear layer ($\mathit{bl}$), which has batch normalization, dropout and ReLu activation. Finally, the output of layer $bl$ is fed into a fully connected output layer with $\mathit{softmax}$ activation that yields the class prediction.
\subsection{Experimental setup}
We extract the IF-THEN rules of the Neuro-Fuzzy model with the methodology of Section~\ref{sec:extract} using 75\% of table $A$ (minus the `MMSI' field) and train a Faster R-CNN model, with `ResNet-50'~\cite{he2015deep} network as backbone using the corresponding images. Then, we extract the deep feature corresponding to each image using the methodology presented in~\ref{sec:extract} and create the datasets $\mathit{IC}$ and $\mathit{VC}$.  Finally, we train both the Neuro-Fuzzy model presented in Section~\ref{sec:methodology} and the baseline model on the created datasets $\mathit{IC}$ and $\mathit{VC}$ and evaluate them separately. 

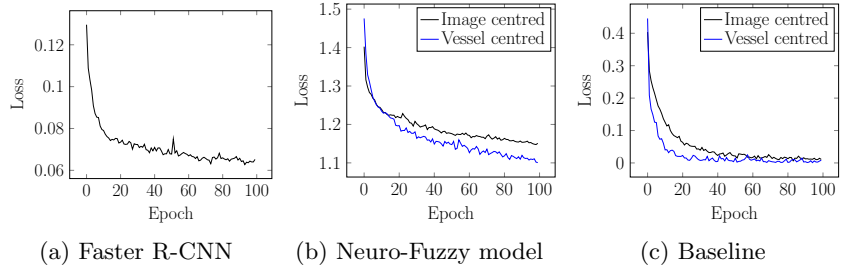
\begin{figure}[t]
\centering
\subfloat[t][Faster R-CNN]{
\resizebox{0.3\textwidth}{!}{
\begin{tikzpicture}
\LARGE
\begin{axis}[xlabel= Epoch,
						 ylabel=Loss,
						 yticklabel style={
        						/pgf/number format/fixed,
        						/pgf/number format/precision=5
						}
]
\addplot[color=black] table [x=epoch,
						 y=loss,
						 col sep=comma] {frcnn_loss.csv};
\end{axis}
\end{tikzpicture}
}}
\subfloat[][Neuro-Fuzzy model]{
\resizebox{0.3\textwidth}{!}{
\begin{tikzpicture}
\LARGE
\begin{axis}[xlabel=Epoch,
						 ylabel=Loss,
						 yticklabel style={
        						/pgf/number format/fixed,
        						/pgf/number format/precision=5
						}
]
\addplot[color=black] table [x=epoch,
						 y=loss,
						 col sep=comma] {nf_loss_ic.csv};
\addplot[color=blue] table [x=epoch,
						 y=loss,
						 col sep=comma] {nf_loss_vc.csv};
\addlegendentry{Image centred}
\addlegendentry{Vessel centred}
\end{axis}
\end{tikzpicture}
}}
\subfloat[][Baseline]{
\resizebox{0.3\textwidth}{!}{
\begin{tikzpicture}
\LARGE
\begin{axis}[xlabel=Epoch,
						 ylabel=Loss,
						 yticklabel style={
        						/pgf/number format/fixed,
        						/pgf/number format/precision=5
						}
]
\addplot[color=black] table [x=epoch,
						 y=loss,
						 col sep=comma] {baseline_loss_ic.csv};
\addplot[color=blue] table [x=epoch,
						 y=loss,
						 col sep=comma] {baseline_loss_vc.csv};
\addlegendentry{Image centred}
\addlegendentry{Vessel centred}
\end{axis}
\end{tikzpicture}
}}

\caption{Plots of the Faster R-CNN (a), the Neuro-Fuzzy model (b) and the baseline (c) losses per epoch number on the Image and Vessel centred (when applicable) training datasets.}
\label{fig:losses}
\end{figure}

All models were implemented using PyTorch~\cite{paszke2017automatic} and were trained for 100 epochs using the Adam optimiser~\cite{kingma2014adam} and with learning rate set to \num{1e-4}. The weight decay for Faster R-CNN was set to to \num{5e-4}, while for the remaining models weight decay was not applied. The batch size was set to 4 and 32 when training the Faster R-CNN model and the Neuro-Fuzzy/Baseline models respectively. Plots of the losses per model are illustrated in Figure~\ref{fig:losses}.
\subsection{Experimental results}
\begin{figure}[ht]
\centering
\subfloat[][Cargo.]{
\includegraphics[width=0.25\textwidth]{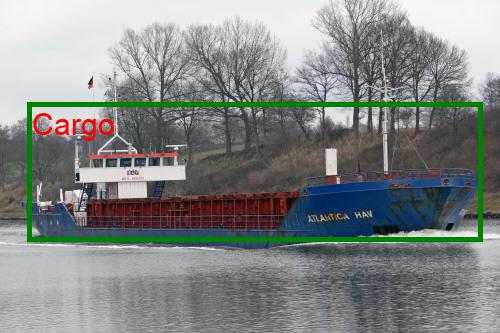}}
\hfill
\subfloat[][Tanker.]{
\includegraphics[width=0.25\textwidth]{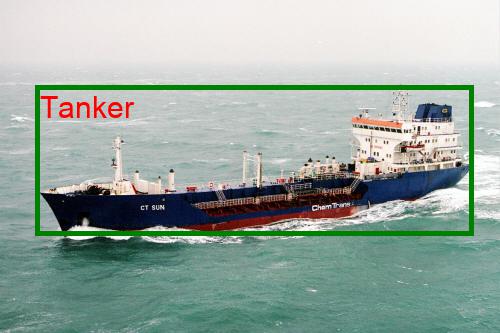}}\hfill
\subfloat[][Tug.]{
\includegraphics[width=0.25\textwidth]{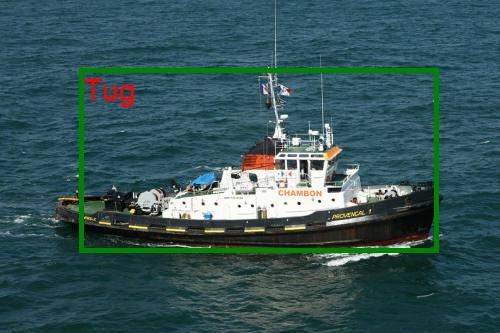}}

\subfloat[][Passenger.]{
\includegraphics[width=0.25\textwidth]{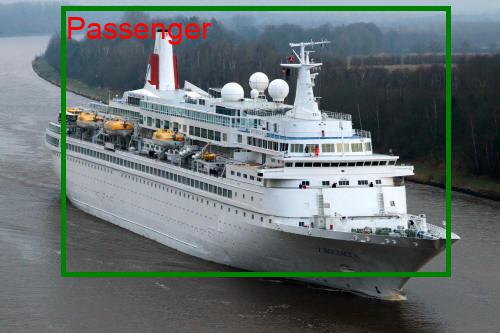}}\hspace{1.7cm}
\subfloat[][Other.]{
\includegraphics[width=0.25\textwidth]{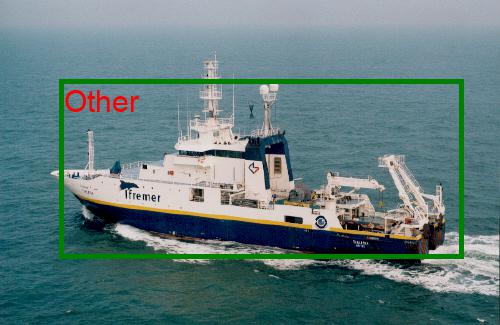}}
\caption{Example detections. The bounding boxes are produced by the Faster R-CNN network while the detected labels are produced from the Neuro-Fuzzy model of this paper.}
\label{fig:example-det}
\end{figure}

We evaluate our model on both Image and Vessel centred datasets. In the first case we use the mean Average Precision Metric presented in \cite{padillaCITE2020}, with interpolation over all recall levels while in the second case we use the macro F1-Score. Some example detections of the Neuro-Fuzzy model are illustated in Figure~\ref{fig:example-det}.
The results of our evaluation are presented in Table~\ref{tab:all-scores} and show that the combination of vessel transmitted AIS information along with Imagery using the Neuro-Fuzzy model of this paper yields better results than using each data source separately and using both sources in the baseline model in both Image and Vessel centred datasets.
\begin{table}[t]
\centering
\caption{mAP scores of the evaluated models. OM, FRCNN, B, DT, kNN, NB, LR, LDA and SVM stand for `Our Model', `Faster R-CNN', `Baseline', `Decision Tree', `k Nearest Neighbours', `Logistic Regression', `Linear Discriminant Analysis' and `Support Vector Machines'. Bold values indicate the highest score per ship and dataset type. The confidence threshold of retaining a bounding box, during the prediction phase of Faster R-CNN, has been set  has been set to $0.7$.
Models with an `*'  used only one source of information i.e., either Images (Image centred) or AIS records (Vessel centred). }
\label{tab:all-scores}
\begin{tabular}{@{}lccc|cccccccc@{}}
\toprule
          & \multicolumn{3}{c}{Image Centred (mAP)}                 & \multicolumn{8}{c}{Vessel Centred (Macro F1-Score)}                                                         \\ \midrule
          & \multicolumn{1}{l}{OM} & \multicolumn{1}{l}{B} & FRCNN* & \multicolumn{1}{l}{OM} & \multicolumn{1}{l}{B} & \multicolumn{1}{l}{DT*} & kNN* & NB*  & LR*  & LDA* & SVM* \\\midrule
Cargo     & 91                     & 92                    & \bblue{94}     & \bred{97}                     & \bred{97}                    & 88                      & 91   & 74   & 86   & 84   & 82   \\

Tanker    & 87                     & 84                    & \bblue{93}     & 94                     & \bred{95}                    & 72                      & 71   & 46   & 46   & 38   & 10   \\
Other     & \bblue{9}                      & \bblue{9}                     & \bblue{9}      & \bred{37}                     & 23                    & 27                      & 30   & 0    & 22   & 0    & 0    \\
Passenger & 94                     & \bblue{97}                    & 84     & 88                     & \bred{94}                    & 37                      & 22   & 46   & 0    & 28   & 0    \\
Tug       & \bblue{67}                     & 59                    & 60     & 88                     & 18                    & 80                      & \bred{89}   & 82   & 40   & 22   & 88   \\

\midrule
All       & \bblue{69.6}          & 68.2                  & 68.0   & \bred{80.8}          & 65.4                  & 60.8                    & 60.6 & 49.6 & 38.8 & 34.4 & 36.0 \\ \bottomrule
\end{tabular}
\end{table}
However, although data fusion proves to improve prediction scores, we attribute the low prediction scores to the class imbalance of the dataset, since the lowest mAP and F1 scores where produced by the `Other' ship type which expresses a diverse spectrum of vessels but has very few examples in the present case. Moreover, in the image centred dataset, although the score difference is not that significant, our model compared to the other two offers to some degree explainability since a classification decision can be tracked through the rules included in the Neuro-Fuzzy system. An example of an extracted rule and its corresponding fuzzified version used for the classification of the `Tug' shiptype is presented in Table~\ref{tab:rule_example}. Table~\ref{tab:rule_example} shows that the fuzzified version of condition~(e) has the highest weight in the disjunction approximation, while also having the highest accuracy in its crisp form on the full dataset, the same behaviour is observed for condition (d) which has the second biggest weight and similarly the second highest accuracy in the full dataset. The remaining weights although they don't match in order their accuracy on the full dataset, they still manage to generalise and reflect a more accurate view of the whole dataset rather than the training dataset.
\begin{table}[t]
 \caption{Example of an extracted logic rule for the `tug' shiptype and its fuzzified version included in the trained Neuro-Fuzzy system. `l', `w', `d', `te', `to' and `tb' stand for the `length', `width', `draught', `to stern', `to port' and `to bow' fields of the AIS messages. The first two columns of the upper part of the table correspond to the rule accuracy, i.e., true positive instances  over total instances covered by the rule on the training and full versions of the vessel centred dataset.}
    \label{tab:rule_example}
    \centering
    \small
    \begin{tabular}{llllc}
    \toprule
    \multicolumn{4}{c}{Extracted rule}\\ 
    Train & Full  & \multicolumn{1}{l}{Conditions} & & Symbol\\\midrule
    &&$\mathit{IF}$&\\
    1 & 0.26 &$\ (\mathit{l} \leq 27.5\ \wedge\ \mathit{te} \leq 16.0\ \wedge\ \mathit{d} > 3.75\ \wedge\ \mathit{to} > 3.5)\ \vee$&&(a)\\[2.5pt]
   0.9 & 0.54 &$\ (\mathit{l} \leq 57.5\ \wedge\ \mathit{w} \leq 24.0\ \wedge\ \mathit{d} > 3.75\ \wedge\ \mathit{l} > 27.5)\ \vee$&&(b)\\[2.5pt]
    1 & 0.5 & $\ (\mathit{l} \leq 72.5\ \wedge\ \mathit{l} > 57.5\ \wedge\ \mathit{te} > 52.0\ \wedge\ \mathit{d} > 5.0)\ \vee$&&(c)\\[2.5pt]
    1 & 0.57&$\ (\mathit{l} \leq 76.5\ \wedge\ \mathit{l} > 72.5\ \wedge \mathit{te} > 52.0)\ \vee$&&(d)\\[2.5pt]
    1&1&$\ (\mathit{tb} \leq 32.5\ \wedge\ \mathit{te} \leq 106.5\ \wedge\ \mathit{l} > 76.5\ \wedge\ \mathit{tb} > 31.5)$&&(e)\\[2.5pt]
    &&$\mathit{THEN}\ \mathit{Tug}$\\\midrule
    
    \multicolumn{4}{c}{Fuzzified rule}\\
    &&\multicolumn{2}{l}{Formula}& Symbol\\\midrule
    \multicolumn{2}{l}{\multirow{10}{*}{$R_{\mathit{Tug}} = \max'\Vast\{$}}&$\ 0.1799\ \min'\lbrace f_{\leq}(\mathit{l};s_{11},27.5),f_{\leq}(\mathit{te};s_{12},16.0),$&\multirow{10}{*}{$\Vast\}$}&\multirow{2}{*}{(a)}\\
    &&$\qquad\qquad\qquad\quad f_{>}(\mathit{d};s_{13},3.75),f_{>}(\mathit{to};s_{14},3.5)\rbrace, $\\[2.5pt]
    &&$\ 0.1035\ \min'\lbrace f_{\leq}(\mathit{l};s_{21},57.5),f_{\leq}(\mathit{w};s_{22},24.0),$&&\multirow{2}{*}{(b)}\\&&$\qquad\qquad\qquad\quad f_{>}(\mathit{d};s_{23},3.75),f_{>}(\mathit{l};s_{24},27.5)\rbrace, $\\[2.5pt]
    &&$\ 0.1089\ \min'\lbrace f_{\leq}(l;s_{31},72.5),f_{>}(l;s_{32},57.5),$&&\multirow{2}{*}{(c)}\\&&$\qquad\qquad\qquad\quad f_{>}(te;s_{33},52.0),f_{>}(d;s_{34},5.0)\rbrace, $\\[2.5pt]
    &&$\ 0.2242\ \min'\lbrace f_{leq}(l;s_{41},76.5),f_{>}(l;s_{42},72.5),$&&\multirow{2}{*}{(d)}\\&&$\qquad\qquad\qquad\quad f_{>}(te;s_{43},52.0)\rbrace, $\\[2.5pt]
    &&$\ 0.3835\ \min'\lbrace f_{\leq}(tb;s_{51};32.5),f_{\leq}(te;s_{52},106.5),$&&\multirow{2}{*}{(e)}\\&&$\qquad\qquad\qquad\quad f_{>}(l;s_{53},76.5),f_{>}(tb;s_{54},31.5)\rbrace$\\\bottomrule
    \end{tabular}
\end{table}

\subsection{Ablation study}
In this section we present our ablation study on the Neuro-Fuzzy model. Our ablation study evolves around the importance of the max depth ($D$) parameter of the CART models used for the rule extraction, and the different values of $r$ used in the weighted exponential means approximation for conjunction and disjunction. We set the $D$ parameter for all trees to 10 and we gradually decrease it in order to limit and in most case decrease the length of the rules. Moreover, for each $D$ setting we train the Neuro-Fuzzy model using different levels for andness and orness ($r$). The results, presented in Table~\ref{tab:ab-results}, show that small rules produced by $D=4$ setting tend to give lower scores. On the other side, over fitted rules (max height set to 10) produce high scores but sacrifice performance due to the increase of parameters. The best score was produced using a depth set to 6. The study shows that `medium high' levels of andness and orness, i.e., $r=-2.14,+2.14$ produce worse results than the `high' and `very high' settings. This is because in the `medium high' setting the model tends to over fit on the training data thus yielding the best training accuracy but loses generalisation. In the `very high' setting the model produces better results however the best score is produced using the `high' setting.
\begin{table}[t]
\caption{F1-Macro scores of the Neuro Fuzzy model produced on the validation set of the Vessel Centred dataset using different parameters for $r$ and $D$. The lower part of the table contains the number of comparisons and conditions included in the extracted rules of each model.}
\label{tab:ab-results}
\centering
\begin{tabular}{l|cccc}
\toprule
  & $D=4$  & $D=6$  & $D=8$  & $D=10$ \\ \midrule
very high \hfill $r=14$   & 72 & 74 & 73 & 76 \\
high \hfill $r=5.4$  & 74 & 81 & 78 & 77 \\
medium high \hfill $r=2.14$ & 69 & 71 & 72 & 76 \\\midrule
\# comparisons & 60 & 264 & 607 & 1126\\
\# conditions & 18 & 58 & 110 & 178\\
\bottomrule
\end{tabular}
\end{table}

%% file: conclusion.tex
\section{Conclusions and Future Work}
\label{sec:sum}
We presented a methodology that can be used to combine AIS data with vessel Imagery along with the advantages of our Neuro-Fuzzy model over the baseline model and using each data source separately as show cased by the experimental evaluation. We believe that the logic rules extracted by the decision trees add information over the dependencies between the AIS fields, and thus providing additional information in the combined Neuro-Fuzzy model. Although our methodology has been applied in the maritime domain, we believe that it can be also applied  in other domains where multiple sources of information are available. 

For future work, we aim to evaluate our Neuro-Fuzzy approach when there is uncertainty or missing fields in the input data, while also investigating further the explainability of our model. Moreover, we want to extend our methodology, so that it handles multiple vessels in an image by automatically linking the AIS transmitted information with the corresponding vessel in the image.